\documentclass[letterpaper]{article}
\usepackage[preprint]{aaai2027}
\usepackage[hyphens]{url}
\usepackage{graphicx}
\usepackage{natbib}
\usepackage{caption}
\usepackage{booktabs}
\usepackage{amsmath}

\title{
VARM-Bench: Benchmarking Verifiable Structured \\ Reasoning in Chinese Abusive Speech Moderation
}

\author{
    Mingyu Yuan\textsuperscript{\rm 1}\equalcontrib,
    Shengtao Wen\textsuperscript{\rm 1}\equalcontrib,
    Lingbing Guo\textsuperscript{\rm 2},
    Zhen Bi\textsuperscript{\rm 3},
    Xiang Chen\textsuperscript{\rm 1}\thanks{Corresponding author.}
}

\affiliations{
    \textsuperscript{\rm 1}MIIT Key Laboratory of Pattern Analysis and Machine Intelligence,\\
College of Computer Science and Technology,\\
Nanjing University of Aeronautics and Astronautics\\
    \textsuperscript{\rm 2}Nanjing University\\
    \textsuperscript{\rm 3}College of Computer Science, Huzhou Normal University\\
    \{xiang\_chen\}@nuaa.edu.cn
}

\newcommand{\dataset}{VARM-Bench}

\begin{document}

\maketitle

\begin{abstract}
The widespread circulation of abusive online content has increased the need for reliable moderation of Chinese social-media text. Existing Chinese benchmarks support label classification, fine-grained toxicity categorization, and target-aware extraction, but do not provide a unified representation for deterministically verifying the stated basis of a moderation decision. We introduce \textbf{VARM-Bench}, a benchmark for field-anchored chain-of-thought rationales in Chinese abusive-speech moderation. Each instance contains a concise natural-language rationale with explicit anchors for six decisions: target, target type, target explicitness, author stance, harmfulness label, and fine-grained category. Our deterministic protocol evaluates field correctness, target alignment, output validity, complete-record agreement, and hidden record errors conditioned on correct final decisions, without relying on an LLM judge. Under a common structured-output protocol, we evaluate language models across multiple model families using zero-shot prompting, taxonomy guidance, and structured CoT supervision, and analyze lexical-cue sensitivity and field-level errors. Results show that strong label-level performance can conceal substantial errors in complete moderation records. VARM-Bench provides an auditable and reproducible benchmark for evaluating verifiable moderation rationales in Chinese abusive-speech moderation.
\end{abstract}

\begin{links}
\link{Code}{https://github.com/NUAA-MMMI/VARM-Bench}
\end{links}
\begin{links}
\link{Dateset}{https://huggingface.co/datasets/NUAA-MMMI/VARM-Bench}
\end{links}

\noindent\textit{\textbf{Content warning:} This paper includes examples containing language that some readers may find offensive or vulgar.}

\section{Introduction}

Abusive-language detection is commonly evaluated using
final-label accuracy~\cite{founta2018large}. A correct label,
however, can conceal an incorrect basis for the moderation
decision. A model may identify the wrong referent, mistake
quoted or rejected abuse as the author's own position, or
assign a harm category that is inconsistent with the actual
target and discourse context while still producing the correct
final label. These errors are particularly consequential in
Chinese social-media text, where implicit references,
homophonic substitutions, negation,
quotation, and sarcasm can change who is being discussed,
and whether apparently
offensive wording is abusive in context.

\begin{figure*}[t]
  \centering
  \includegraphics[width=0.95\linewidth]{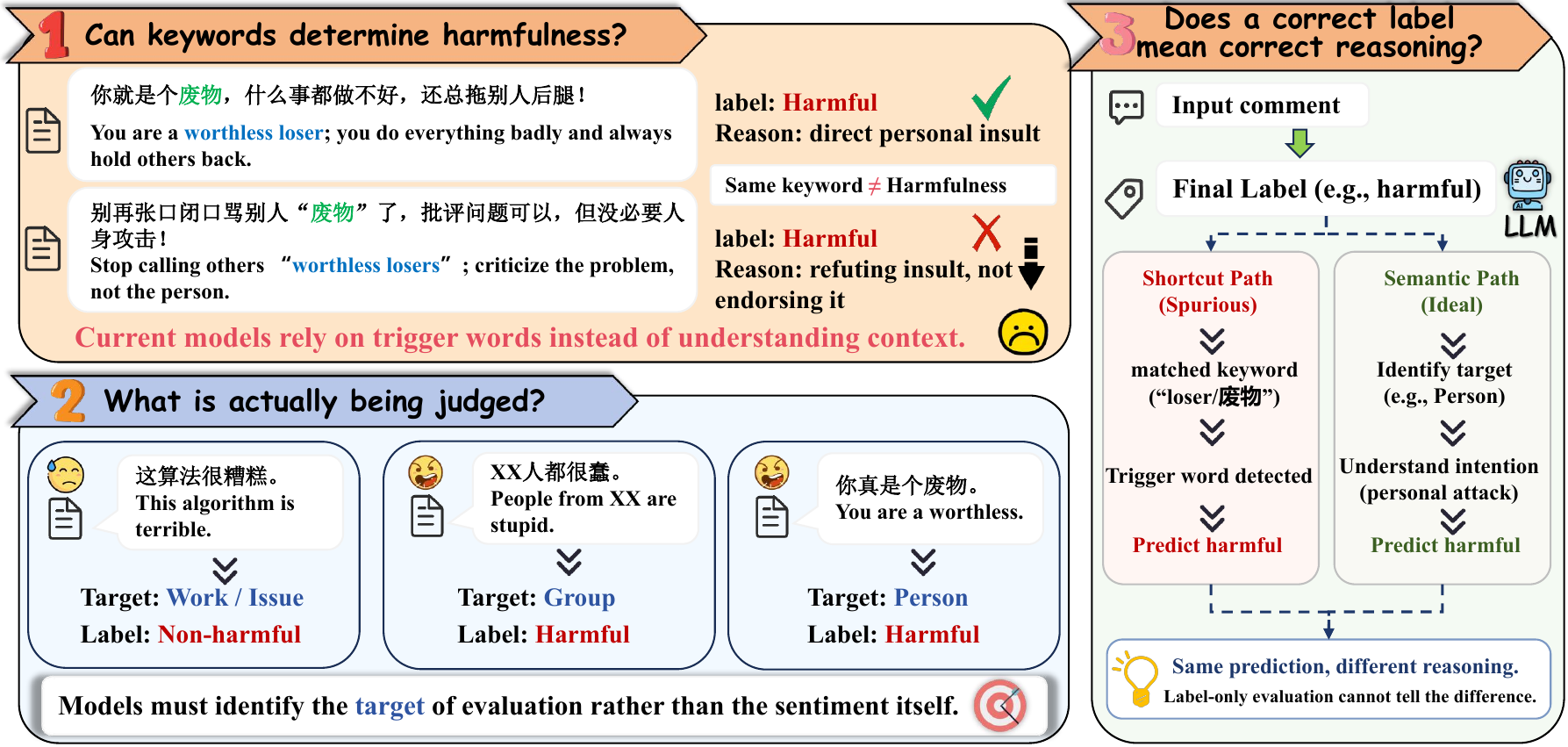}
  \caption{Motivation for \dataset{}. Similar lexical cues can
  correspond to attacks, quotations, opposition, or neutral
  mentions. A final label alone cannot reveal whether the target,
  stance, and harm category are jointly correct.}
  \label{fig:motivation}
\end{figure*}

Existing Chinese benchmarks have advanced from offensive-language classification and social-bias detection to implicit-toxicity analysis, fine-grained categorization, and target-aware structured extraction~\cite{deng2022cold,jiang2022swsr}. Nevertheless, labels, categories, and target spans represent only isolated components of a moderation decision. A target span does not reveal whether the author attacks, rejects, quotes, or neutrally mentions the target. Conversely, a structured field tuple does not explain which input cues support its predictions. Free-form rationales can express these relations, but their central claims are difficult to extract and evaluate consistently. As shown in Figure~\ref{fig:motivation},  reliable
evaluation therefore requires a unified representation in which
the stated reasoning is expressed in natural language, its key
moderation decisions are explicitly recoverable, and the
resulting record can be checked deterministically.

We introduce \dataset{} (Verifiable Abusive Reasoning Moderation Benchmark), for field-anchored chain-of-thought (CoT) in Chinese abusive-speech moderation. For each input, a model generates one explanation with explicit anchors for six decisions: target, target type, target explicitness, author stance, harmfulness, and fine-grained category. A deterministic parser converts these anchors into a complete moderation record, making every scored decision traceable to the explanation and verifiable against the input. We treat the generated CoT only as an inspectable rationale, not as a faithful account of the model’s internal reasoning.

\dataset{} contains 8{,}000 Chinese social-media comments, including 1{,}440 challenging non-harmful examples involving quotation, opposition, neutral identity mentions, behavioral criticism, and other misleading surface cues. After model-assisted pre-annotation, every reference record was manually corrected and rechecked against a shared codebook. Evaluation covers final-label and category performance, as well as deterministic agreement between records extracted from output CoTs and their references. Separate audits assess sensitivity to reference conventions and explanation quality.

We establish a unified protocol for generating, recovering, and evaluating complete moderation records. We use this protocol to benchmark multiple open and closed models across prompting and supervision settings and provide diagnostic analyses of hidden record errors, context-dependent lexical cues, field-level bottlenecks, and generated-rationale quality. Together, the benchmark, protocol, and analyses provide a reproducible foundation for studying verifiable reasoning in Chinese abusive-speech moderation. Our main contributions are summarized as follows:
\begin{itemize}
\item We introduce a Chinese abusive-speech benchmark in which six moderation decisions are embedded in one natural-language CoT and deterministically reconstructed as a complete moderation record.
\item We establish a multi-level evaluation protocol separating decision quality, field correctness, record validity, and complete-record agreement, with human review and audits of references and rationales.
\item We evaluate open and closed models on VARM-Bench across prompting and supervision settings and analyze hidden record errors, lexical-cue sensitivity, and field bottlenecks, identifying referent localization as the main bottleneck under frozen-reference scoring.
\end{itemize}

\section{Related Work}

\paragraph{Abusive-Language Detection and Chinese Benchmarks.}
Early abusive-language datasets mainly evaluated post-level classification~\cite{DBLP:conf/icwsm/DavidsonWMW17}. Later work broadened evaluation to implicit toxicity and target-based offensive-language identification~\cite{DBLP:conf/acl/HartvigsenGPSRK22,zampieri2023target}. Toxic-span benchmarks and functional tests further examine localized evidence and behavioral failures in hate-speech systems~\cite{pavlopoulos2021semeval,rottger-etal-2021-hatecheck}. Chinese benchmarks make parallel progress. COLD and SWSR cover offensive language and online sexism~\cite{deng2022cold,jiang2022swsr}; CDial-Bias and ToxiCN add targeted groups, implied attitudes, toxicity types, and expression forms~\cite{zhou2022identifyingsocialbiasdialog,lu2023toxicn}. STATE-ToxiCN moves to span-level Target--Argument--Hateful--Group extraction, while ChineseHarm-Bench covers policy-violation categories with an expert-curated rule base~\cite{bai2025statetoxicn,liu2025chineseharmbench}. Contextual resources such as Social Bias Frames and Latent Hatred show why intent, social implication, and implicit-hate explanations matter beyond surface toxicity~\cite{sap2020socialbiasframes,DBLP:conf/emnlp/ElSheriefZMASCY21}. Moderation studies further show that decisions vary with annotation definitions and conversational context~\cite{schopke2023defineharm,yu2022counterspeech}. Policy goals and moderator needs also affect how similar cues should be handled~\cite{cao2024toxicitynotall,zheng2023hatemoderate}. This work motivates evaluation beyond a final label or isolated span: the target, its properties, author stance, harmfulness, and category must be checked together to determine whether a moderation decision rests on the right basis. Existing benchmarks provide labels, taxonomies, target annotations, and structured extraction, but none reconstructs this complete moderation record from one generated rationale.

\begin{figure*}[t]
  \centering
  \includegraphics[width=0.95\linewidth]{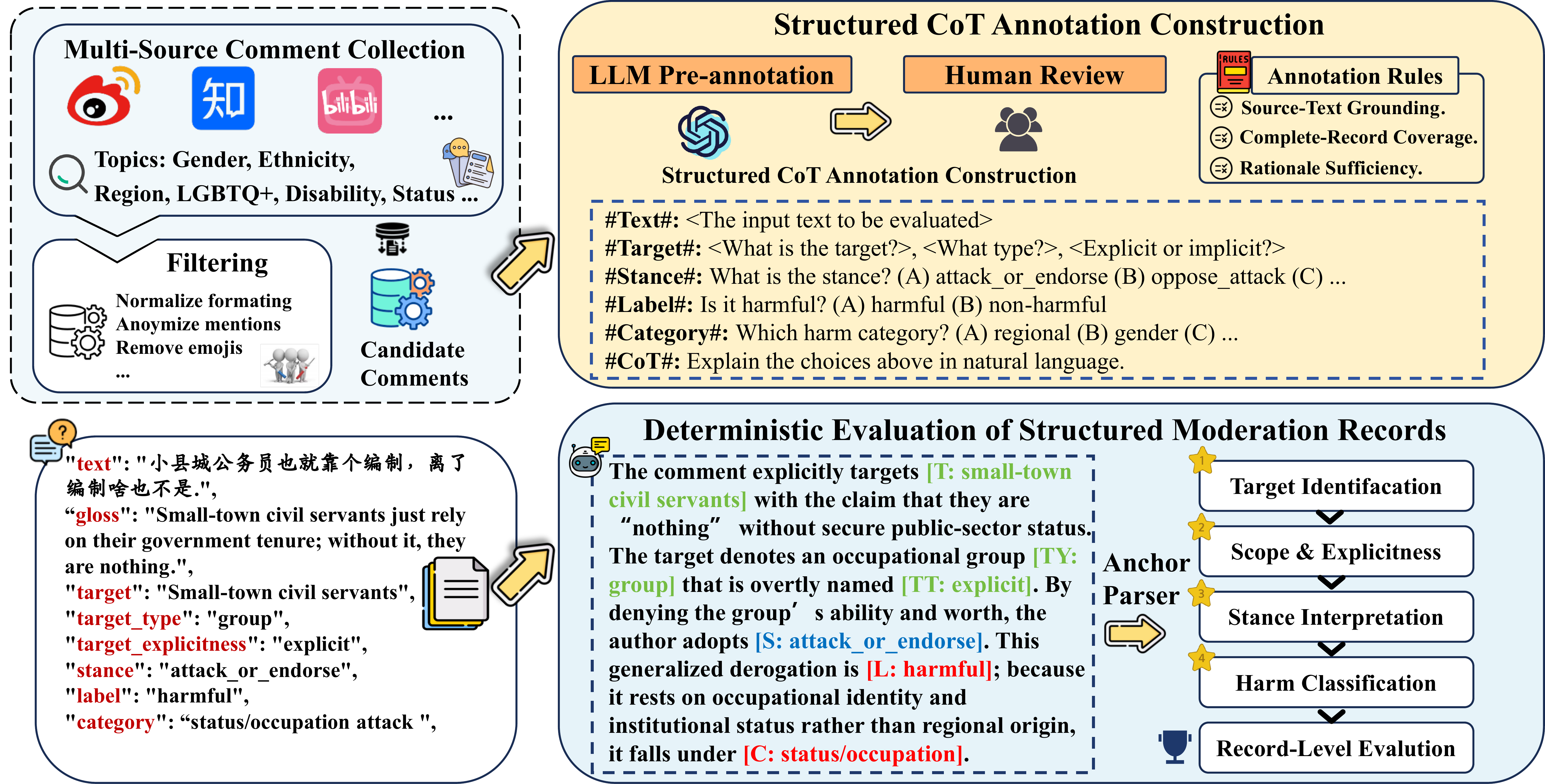}
  \caption{Overview of the data construction, annotation, and validation pipeline. Multi-source Chinese comments are normalized, anonymized, and filtered, annotated through LLM-assisted pre-annotation and human review, and finally evaluated at the target, scope, stance, harm, and record levels.}
  \label{fig:framework}
\end{figure*}

\paragraph{Rationales and Structured Verification.}
Rationale benchmarks such as HateXplain and ERASER provide human-annotated evidence for model predictions~\cite{mathew2021hatexplain,deyoung2020eraser}. Such evidence makes predictions easier to inspect, but extracted spans do not show how the target, author stance, harmfulness label, and fine-grained category jointly support a moderation decision. Chain-of-thought prompting made natural-language reasoning a common prediction interface~\cite{wei2022chain}. Self-consistency showed that sampling multiple reasoning paths can improve prediction reliability~\cite{wang2023self}. Later work evaluated intermediate reasoning with chain-level metrics and process supervision~\cite{golovneva2022roscoe,prasad2023receval}. Stepwise verification further formalized process-level supervision for complex reasoning~\cite{lightman2023verify}. Model-based evaluators extended automatic assessment to broader judgment tasks~\cite{zheng2023judgingllmasajudgemtbenchchatbot}. Safety benchmarks test complementary properties such as safety knowledge, trustworthiness, adversarial robustness, and refusal behavior~\cite{zhang2024safetybenchevaluatingsafetylarge,wang2023decodingtrust}. HarmBench focuses more directly on automated red teaming and harmful-behavior evaluation~\cite{mazeika2024harmbench}. Natural-language rationales can express relations among moderation fields, but their core judgments cannot be extracted and compared consistently without explicit checkpoints. Generated CoT explanations may also diverge from the factors behind predictions~\cite{lanham2023measuringfaithfulnesschainofthoughtreasoning,turpin2023languagemodelsdontsay}. In \dataset{}, each natural-language CoT contains six checkpoints. A deterministic parser reconstructs the complete record, and a manual audit evaluates grounding, contextual correctness, and support for the predicted decision.

\section{Preliminaries}

\subsection{Motivation: Why Verifiable Structured Reasoning?}

Final-label evaluation reveals a moderation decision but not whether that decision follows from a correct interpretation of the input. A model may return the correct label while identifying the wrong target, misinterpreting target explicitness or author stance, or assigning an incompatible category. Field tuples support deterministic scoring but omit input-specific justification. Free-form rationales capture contextual relations, but they are difficult to extract and compare consistently. VARM-Bench combines these functions in a single natural-language moderation CoT with six explicit anchors for target, target type, target explicitness, author stance, harmfulness label, and fine-grained category. A deterministic parser reconstructs the stated record for field-level and joint scoring, while the surrounding rationale enables audits of grounding, contextual correctness, and inferential sufficiency. In this paper, verifiable reasoning refers only to this operational property. It does not imply access to \mbox{latent model computation}.

\subsection{Task Definition}

Each input $x_i$ is paired with a six-field reference moderation record
\begin{equation}
z_i =
(t_i,\tau_i,\epsilon_i,s_i,y_i,c_i),
\end{equation}
and a field-anchored CoT $r_i$ that justifies its decisions. The benchmark is
\begin{equation}
\mathcal{D}
=
\left\{
(x_i,z_i,r_i)
\right\}_{i=1}^{N}.
\end{equation}

Given only the input text $x_i$, the model generates a single natural-language moderation CoT:
\begin{equation}
\hat r_i = f_{\theta}(x_i).
\end{equation}
The generated CoT uses this fixed sequence of six anchors:
\[
\texttt{[T:]}
\rightarrow
\texttt{[TY:]}
\rightarrow
\texttt{[TT:]}
\rightarrow
\texttt{[S:]}
\rightarrow
\texttt{[L:]}
\rightarrow
\texttt{[C:]},
\]
which encode the target, target type, target explicitness, author stance, harmfulness label, and fine-grained category, respectively. A deterministic parser $\mathcal{P}$ extracts the anchored values and reconstructs the predicted moderation record:
\begin{equation}
\hat z_i
=
\mathcal{P}(\hat r_i)
=
(\hat t_i,\hat\tau_i,\hat\epsilon_i,
 \hat s_i,\hat y_i,\hat c_i).
\end{equation}
No separate field tuple is generated: all field-level predictions are parsed directly from the same generated CoT, whose surrounding text is retained for rationale auditing. The six parsed fields, together with their operational definitions and admissible values, are specified as follows:

\begin{itemize}
    \item \textbf{Target} ($t_i$) identifies the shortest stable
    referent of the main decision-relevant proposition. Incidental
    entities are excluded, while coordinated referents under the same
    proposition are treated as one target. In quoted or opposed abuse,
    the target remains the referent of the embedded claim.
    \emph{Output:} a normalized free-text referent or
    \textit{no clear target}.

    \item \textbf{Target Type} ($\tau_i$) describes the referential
    scope of the target. \emph{Values:} \textit{single object},
    \textit{group object}, and \textit{no clear target}.

    \item \textbf{Target Explicitness} ($\epsilon_i$) indicates whether
    the target is directly expressed or inferred from context.
    \emph{Values:} \textit{explicit}, \textit{implicit}, and
    \textit{no clear target}.

    \item \textbf{Stance} ($s_i$) describes the author's relation to
    the relevant abusive claim. \emph{Values:}
    \textit{attack or endorse}, \textit{oppose attack},
    \textit{quote or report}, and \textit{neutral mention}.

    \item \textbf{Label} ($y_i$) indicates whether the text performs or
    endorses abuse. \emph{Values:} \textit{harmful} and
    \textit{non-harmful}.

    \item \textbf{Category} ($c_i$) identifies the primary basis of the
    moderation decision. \emph{Values:} \textit{general abuse},
    \textit{region/ethnicity}, \textit{gender},
    \textit{sexual orientation and gender identity (SOGI)},
    \textit{status/occupation}, \textit{body/health}, and
    \textit{non-attack}. All non-harmful instances use
    \textit{non-attack}.
\end{itemize}

Detailed tie-breaking rules appear in the Appendix. Prose between anchors must link input-specific evidence to decisions, especially for context-sensitive phenomena such as quotation, negation or opposition, implicit reference, sarcasm, and behavioral criticism. Listing or paraphrasing anchor values may satisfy the format but not the rationale requirement. Automatic evaluation measures field-level and joint agreement with the reference record.

\section{Benchmark Construction}

Figure~\ref{fig:framework} summarizes the construction pipeline.
\dataset{} contains 8,000 records split into 5,600 training, 800
development, and 1,600 test instances, with a consistent 55/45
harmful--non-harmful ratio and 1,440 difficult non-harmful cases.
To prevent cross-split leakage, normalized duplicates and pairs with
RapidFuzz similarity of at least 95\% are confined to a single split.

\subsection{Data Collection and Filtering}

We collected public posts and comments from Bilibili, Zhihu, Baidu Tieba, and Hupu. Topic-based queries provided candidates for the six harmful categories. Related lexical cues were also used to find difficult non-harmful examples, including quotations of abusive language, objections to abuse, neutral identity references, personal accounts, and criticism of behavior. These examples were labeled as \textit{non-harmful} and assigned to the \textit{non-attack} category. Search terms were used only for retrieval; annotators made the final decisions from the referent, the author's stance, and the full context. We kept self-contained texts of at least 15 characters and discarded advertisements, corrupted text, meaningless symbol strings, and replies that required missing context. We removed URLs and markup and deleted or generalized personal identifiers without changing the intended meaning. Finally, we applied Unicode NFKC normalization, removed exact duplicates, and excluded text pairs with RapidFuzz similarity of 95\% or higher, including cross-split pairs.

\begin{figure}[t]
  \centering
  \includegraphics[width=\columnwidth]{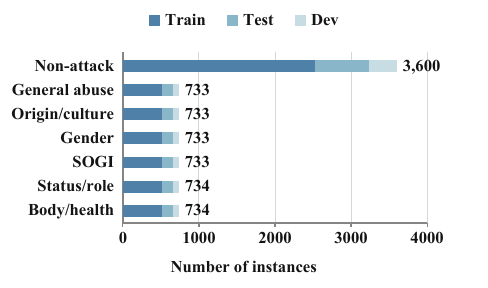}
\caption{Seven-way category distribution of \dataset{} across the train, development, and test splits.}
\label{fig:category}
\end{figure}

\begin{table*}[t]
\centering
\footnotesize
\setlength{\tabcolsep}{2.4pt}
\begin{tabular*}{\textwidth}{@{\extracolsep{\fill}}llccccccccccc@{}}
\toprule
\textbf{Model} & \textbf{Setting}
& \multicolumn{2}{c}{\textbf{Decision}}
& \multicolumn{4}{c}{\textbf{Structured Fields}}
& \multicolumn{2}{c}{\textbf{Validity}}
& \textbf{Joint}
& \multicolumn{2}{c}{\textbf{Hidden Error}} \\
\cmidrule(lr){3-4}
\cmidrule(lr){5-8}
\cmidrule(lr){9-10}
\cmidrule(lr){11-11}
\cmidrule(lr){12-13}
& & \textbf{Label} & \textbf{Cat.}  & \textbf{T-F1} & \textbf{Type} & \textbf{Exp.} & \textbf{Stance} & \textbf{Parse} & \textbf{Format} & \textbf{JREM} & \textbf{HER-C} & \textbf{HER-L} \\

\midrule
Qwen2.5-7B & Zero-shot
& 88.4 & 63.7 & 57.5 & 50.1 & 34.9 & 54.4 & 94.0 & 91.6 & 31.9 & 54.9 & 62.8 \\
Qwen2.5-7B & +Taxonomy
& 90.6 & 76.7 & 58.0 & 48.9 & 36.8 & 53.2 & 97.1 & 94.0 & 36.4 & 54.8 & 59.3 \\
Qwen2.5-7B & CoT-SFT
& 96.7 & \textbf{88.2} & \textbf{73.1} & 59.2 & 46.9 & 78.4 & 99.6 & 99.4 & \textbf{59.5} & \textbf{34.4} & 38.3 \\

\midrule
Llama-3.1-8B & Zero-shot
& 61.4 & 54.8 & 50.0 & 51.0 & 34.1 & 30.6 & 86.6 & 82.1 & 24.6 & 48.0 & 61.6 \\
Llama-3.1-8B & +Taxonomy
& 80.1 & 63.6 & 52.2 & 53.9 & 34.6 & 44.3 & 94.2 & 91.3 & 30.4 & 51.7 & 61.7 \\
Llama-3.1-8B & CoT-SFT
& 95.3 & 86.3 & 69.2 & 61.5 & 46.0 & \textbf{80.9} & 99.5 & 99.3 & 56.0 & 37.2 & 41.1 \\

\midrule
InternLM3-8B & Zero-shot
& 77.6 & 56.7 & 31.8 & 44.4 & 24.3 & 42.5 & 77.8 & 53.3 & 7.9 & 86.3 & 88.4 \\
InternLM3-8B & +Taxonomy
& 79.9 & 63.9 & 36.7 & 47.1 & 25.8 & 52.0 & 83.4 & 72.0 & 12.7 & 79.8 & 82.8 \\
InternLM3-8B & CoT-SFT
& 96.1 & 88.1 & 72.3 & \textbf{65.7} & \textbf{53.7} & 75.4 & \textbf{100.0} & 99.8 & \textbf{59.5} & 34.5 & \textbf{38.1} \\

\midrule
GPT-5.5 & Zero-shot
& \textbf{97.6} & 85.7 & 69.1 & 59.5 & 39.2 & 76.1 & \textbf{100.0} & \textbf{100.0} & 55.4 & 38.1 & 43.2 \\
Qwen3.7-Max & Zero-shot
& 96.9 & 86.4 & 61.7 & 60.7 & 40.6 & 66.6 & 99.9 & 99.4 & 49.6 & 45.0 & 48.9 \\
DeepSeek-V4-Pro & Zero-shot
& 95.6 & 86.8 & 61.1 & 63.0 & 44.0 & 67.0 & 99.0 & 97.4 & 46.3 & 48.1 & 51.4 \\

\bottomrule
\end{tabular*}
\caption{Performance comparison of different models on the deduplicated 1,600-item test set (\%) across classification, target extraction, structural validity, and complete-record evaluation. Classification metrics use Macro-F1, while T-F1 uses mean character-F1. HER-C/L are lower-is-better; all other metrics are higher-is-better. Bold indicates the best result in each column.}
\label{tab:mainresults}
\end{table*}

\subsection{Annotation Schema and Procedure}

LLM-assisted pre-annotation produced a draft six-field record and an 80--180-character anchored CoT. Three master's students and one doctoral student then independently reviewed all 8,000 input--record--rationale triples using a shared codebook. The first pass revised every field and explanatory link; the second verified each record against the source text and flagged unresolved cases. Accepted rationales contained each anchor once, in order. The codebook specified tie-breaking rules for category boundaries, implicit targets, quoted or opposed abuse, and multi-attribute cases. Quoted, opposed, or neutral mentions remained non-harmful unless the author endorsed the attack, while the attribute supporting the main abusive claim determined the primary category. Reviewers revised or rejected unsupported targets, incorrect stance attributions, incompatible label--category assignments, anchor-only restatements, and unsupported input-to-decision links. The final release retained only source-grounded, mutually consistent, human-verified records.

\subsection{Annotation Quality Control.}
A record was accepted only when its target attributes, stance, label, category, and rationale were supported by the source text and mutually consistent. A dataset-wide validator checked completeness, valid values, anchor order and uniqueness, and cross-field constraints. Failed records were returned for human correction. All 8,000 released records passed these checks, and no cross-split duplicates remained at the 95\% RapidFuzz threshold. To assess annotation consistency, three trained reviewers independently reannotated a category- and phenomenon-stratified sample of 200 records. Reviewers saw only the source text and had no access to the frozen reference or each other's decisions. Krippendorff's $\alpha$ was 0.943 for Label, 0.935 for Category, 0.914 for Stance, 0.797 for Target Type, and 0.671 for Target Explicitness. Target is a normalized free-text referent, so its agreement was measured with pairwise character-F1, yielding 0.775. These results show high agreement on core moderation decisions, with greater contextual sensitivity in Target Explicitness. Disagreements were adjudicated after all independent judgments were recorded.

\subsection{Evaluation Metrics}

Metrics are computed end to end: unparseable outputs remain in the denominators and receive zero. We report Macro-F1 for the five discrete fields, normalized character-overlap F1 (T-F1) for the free-text target, and Parse Success when all six valid anchors occur once and in order. JREM follows the frozen reference convention rather than asserting a uniquely correct semantic record.

\begin{figure*}[t]
  \centering
  \includegraphics[width=0.92\textwidth]{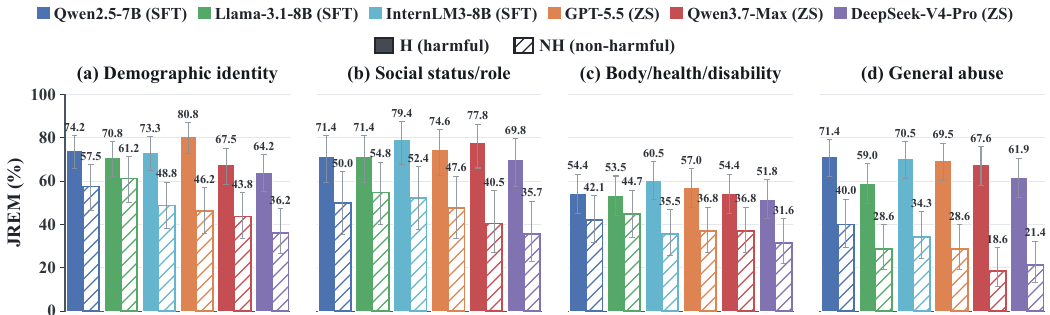}
  \caption{JREM across four context-dependent lexical-cue families. Solid and hatched bars denote harmful (H) and non-harmful (NH) cases; overlapping families are evaluated independently.}
  \label{fig:q2-phenomena}
\end{figure*}

\begin{figure*}[t]
  \centering
  \includegraphics[width=0.92\textwidth]{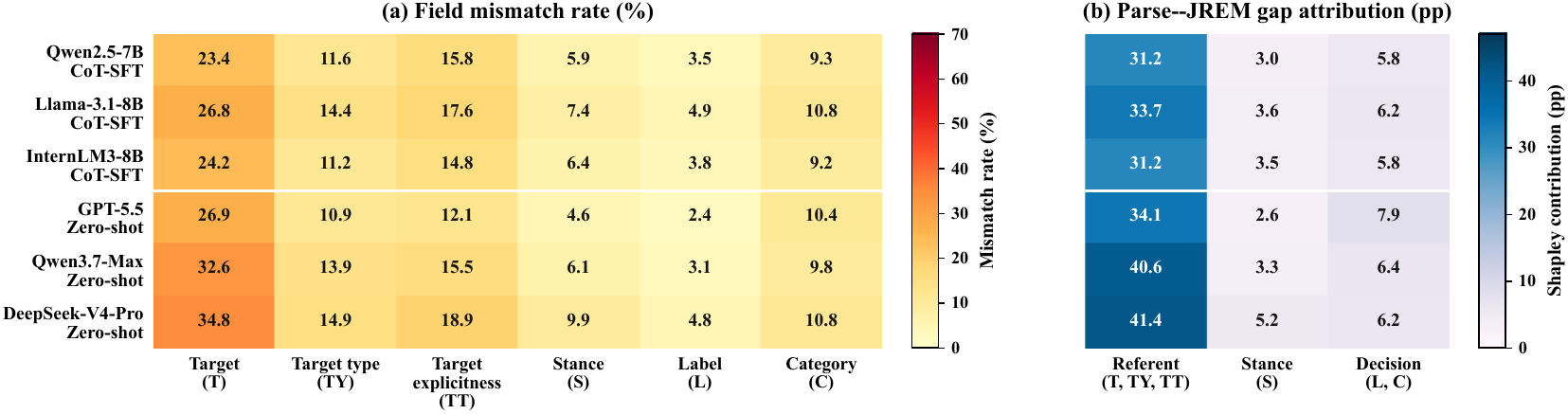}
  \caption{Complete-record error analysis on the 1,600-item test set. (a) Field mismatch rates, with target F1 below 0.5 counted as a mismatch; (b) Shapley decomposition of the Parse--JREM gap across referent, stance, and decision fields.}
  \label{fig:q3-field-bottlenecks}
\end{figure*}
Let $J_i=1$ when output $i$ is parseable, its target T-F1 is at least 0.5, and the other five fields match the reference; otherwise $J_i=0$. Let $M_i^X$ indicate a correct category ($X=C$) or label ($X=L$) prediction:
\begin{gather}
\mathrm{JREM} = \frac{1}{N}\sum_{i=1}^{N}J_i, \\
\mathrm{HER\text{-}X}=
\frac{\sum_{i=1}^{N}M_i^X(1-J_i)}
{\sum_{i=1}^{N}M_i^X},\quad X\in\{C,L\}.
\end{gather}
HER-C and HER-L therefore measure hidden complete-record errors among category- and label-correct outputs; parse failures lower JREM but do not enter their denominators. Higher is better except for HER-C and HER-L. 

\section{Experiments}

\subsection{Experimental Setup}

We evaluate Qwen2.5-7B, Llama-3.1-8B, and InternLM3-8B under zero-shot, taxonomy-guided zero-shot, and CoT-SFT; GPT-5.5, Qwen3.7-Max, and DeepSeek-V4-Pro use zero-shot. All systems share the deduplicated 1,600-item test set. CoT-SFT uses 5,600 training and 800 development instances with prompt-masked LoRA and development-loss early stopping, directly supervising all six anchors and their connecting prose. Across settings, the test inputs, six-anchor output contract, target normalization, parser, and metrics remain fixed; only the prompting or adaptation condition changes. Each table entry is recomputed from its frozen full-test output, with no manual repair of parse or field errors. The test set is excluded from adaptation and model selection. Deterministic decoding and one parser score every output; the Appendix documents prompts, hyperparameters, hardware, and implementation needed for reproduction.

\subsection{Main Results}

\begin{figure*}[t]
  \centering
  \includegraphics[width=0.9\textwidth]{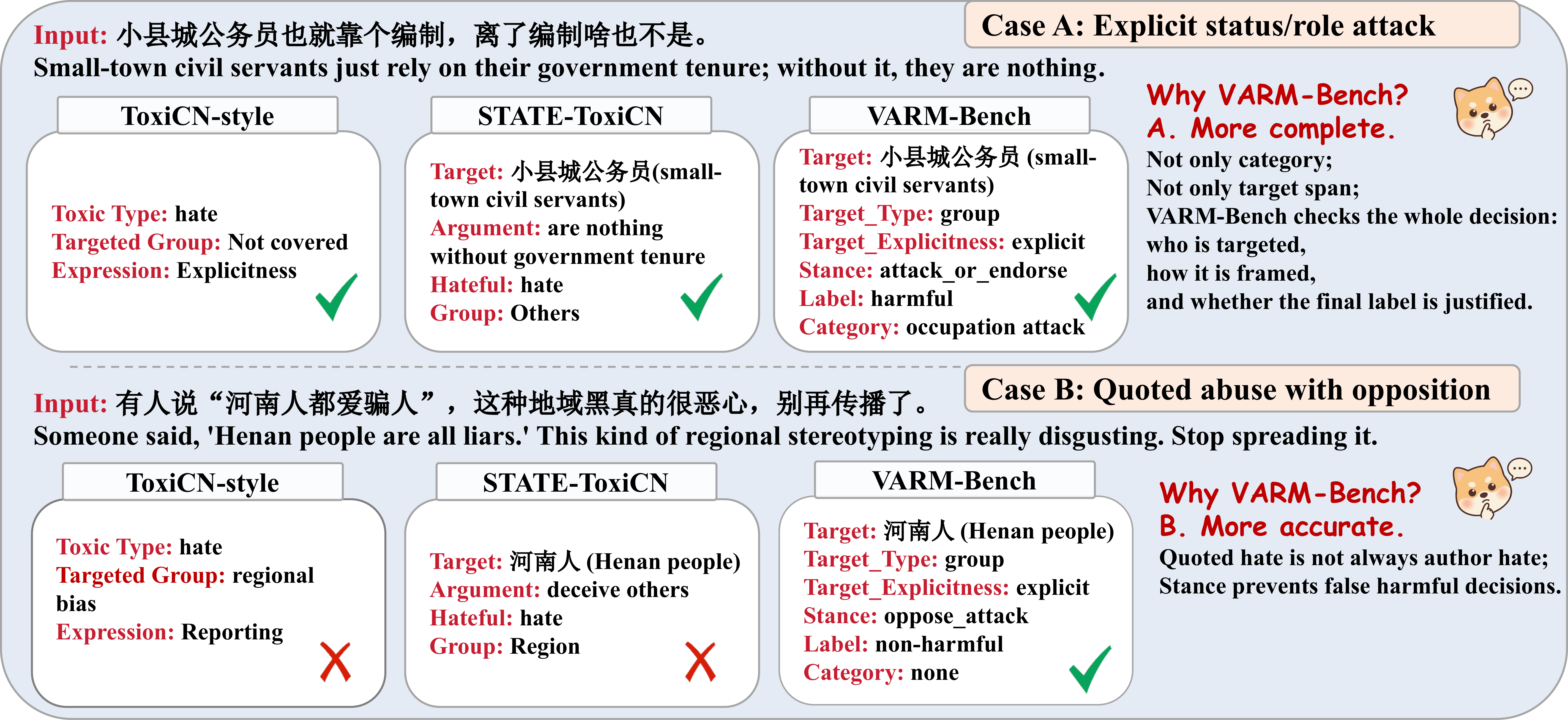}
  \caption{Qualitative comparison of moderation representations. Case A shows that label/category and target-span outputs omit target properties and the linked decision basis. Case B shows that identifying quoted abuse without author stance can invert harmfulness. \dataset{} makes all six decisions jointly inspectable.}
  \label{fig:caseanalysis}
\end{figure*}

\paragraph{Q1: How often do correct labels conceal record errors?}
To assess whether label-level performance reflects complete-record quality, we evaluate all models and training settings on the same 1,600-item test set using JREM and HER-C/HER-L. JREM requires the target to meet the character-overlap threshold and the other five fields to match the reference. HER-C/HER-L measure the record errors that remain among parseable outputs with a correct category or label. Table~\ref{tab:mainresults} shows that GPT-5.5 achieves Label and Category Macro-F1 scores of 97.6\% and 85.7\%, but its JREM is only 55.4\%; 38.1\% of its category-correct outputs still contain a record error. For the three open models, taxonomy guidance raises Category Macro-F1 by 7.2--13.0 points, while HER-C remains at 51.7--79.8\%. CoT-SFT improves the complete record more consistently, raising JREM to 56.0--59.5\% and lowering HER-C to 34.4--37.2\%. Parse Success is nearly 100\% for the CoT-SFT systems, so most remaining errors concern field content. A model may select the correct category while assigning it to the wrong target, explicitness, or author stance. Label-level scores therefore provide only a partial account of complete-record quality.

\paragraph{Q2: How sensitive are models to lexical cues?}
Lexical cues alone do not determine whether a text is harmful, but strongly abusive terms can influence model judgments. We construct four challenge subsets from the test set, covering demographic identity, social status or role, body, health or disability, and general abuse. Each subset contains approximately 100--200 instances with a controlled 60:40 harmful-to-non-harmful ratio. As shown in Figure~\ref{fig:q2-phenomena}, harmful cases achieve higher JREM in all 24 model--subset comparisons, with a median gap of 26.0 percentage points. General-abuse cues produce the largest mean gap at 38.1 points, whereas the body, health, or disability subset has the lowest overall JREM for both harmful and non-harmful cases. CoT-SFT and zero-shot API systems perform similarly on harmful cases, averaging 67.5\% and 66.4\% JREM, respectively. Their non-harmful JREM differs more clearly, reaching 45.8\% for CoT-SFT and 35.3\% for the API systems. The advantage of CoT-SFT is therefore concentrated in quotations, rebuttals, neutral mentions, and other non-harmful uses of salient cues. These results show that surface cues remain particularly difficult when their presence does not express the author's own attack, while structured supervision improves the consistency of the resulting moderation records.

\paragraph{Q3: Which fields limit complete-record prediction?}
Figure~\ref{fig:q3-field-bottlenecks} examines bottlenecks at two complementary levels. At the field level, the target has the highest mismatch rate for every system, followed by target explicitness, whereas the harmfulness label is comparatively stable. This ordering matters because target localization determines the referent whose type, explicitness, stance relation, and category must then be interpreted. A boundary or normalization error can therefore coincide with several downstream mismatches even when the final decision is correct. The exact Shapley decomposition accounts for this overlap rather than counting each field independently; it attributes 76.6--80.6\% of the Parse Success--JREM gap to the Referent group. The field and group analyses thus converge on target localization as the main scored bottleneck. At the same time, the audit panel shows that reasonable readers do not always choose the same normalized target or explicitness value as the frozen reference. The measured gap consequently reflects both model difficulty and a single-answer convention. Better referent resolution, acceptance of equivalent target descriptions, and reviewer correction are therefore more appropriate remedies than treating every disagreement as an unambiguous semantic failure. The rationale audit remains separate from reference labels and JREM.

\begin{figure}[t]
  \centering
  \includegraphics[width=\columnwidth]{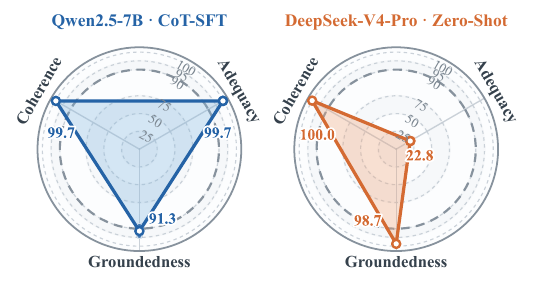}
  \caption{Generated-CoT audit profiles for Qwen2.5-7B CoT-SFT and DeepSeek-V4-Pro Zero-Shot. The three axes show Groundedness, Adequacy, and Coherence across 600 CoTs from 300 shared inputs after three blinded GPT-5.5 passes and human review.}
  \label{fig:generated-cot-quality}
\end{figure}

\paragraph{Q4: Can generated CoTs support human review?}
To assess whether generated CoTs are suitable for human inspection, we randomly sample 300 shared test inputs without replacement from Qwen2.5-7B CoT-SFT and DeepSeek-V4-Pro Zero-Shot, yielding 600 CoTs. We propose three quality dimensions: \textbf{Groundedness} checks whether the rationale introduces source-unsupported claims; \textbf{Adequacy} requires sample-specific justification rather than field restatement; and \textbf{Coherence} tests consistency within the rationale and with the predicted record. Each CoT undergoes three blinded GPT-5.5 review passes, followed by human rechecking of failed, non-unanimous, and sampled unanimous items. CoT-SFT obtains 91.3\%, 99.7\%, and 99.7\% on the three dimensions, whereas Zero-Shot obtains 98.7\%, 22.8\%, and 100.0\% (Figure~\ref{fig:generated-cot-quality}). These results show that grounded and coherent rationales can remain too generic for inspection, while CoT-SFT more consistently provides sample-specific explanations. Generated CoTs can thus serve as inspectable review drafts, with human retaining final authority.

\subsection{Case Study}
Figure~\ref{fig:caseanalysis} shows why complete-record evaluation matters. In Case A, all representations recognize a status-based attack, but label/category output omits the affected group, while target--argument extraction omits explicitness and the link from author stance to the final decision. The \dataset{} record makes these dependencies visible by linking the explicit group target to an attacking stance, a harmful label, and the status/occupation category. Case B exposes a different failure. The text quotes a regional stereotype only to reject it; representations without author stance can treat the embedded claim as the author position and produce a false harmful decision. \dataset{} retains the quoted group as the target but records opposition, non-harmfulness, and no harm category. Plausible fields can still form an inconsistent record. Together, target extraction alone cannot resolve decisions involving speaker commitment or discourse function.

\section{Conclusion and Future Outlook}

In this paper, we present VARM-Bench, a comprehensive benchmark for evaluating field-anchored rationales in Chinese abusive-speech moderation. The benchmark reconstructs complete moderation records from generated CoT rationales and evaluates multiple aspects of moderation decisions, including target identification, target properties, author stance, harmfulness labels, and categories through deterministic metrics. Our results demonstrate that strong label-level performance can coexist with substantial errors in the underlying moderation records, with target identification emerging as the primary source of failure. Our benchmark enables systematic analysis of complete moderation decisions and model failure patterns. Future work will extend VARM-Bench to additional domains and diverse moderation policies, improve the handling of semantically equivalent target descriptions, and investigate cross-lingual transfer. Broader human evaluation can further assess the grounding and contextual validity of generated rationales, providing deeper insights into their reliability in real-world moderation scenarios.

\bibliography{aaai2027}

\end{document}